\documentclass[10pt,twocolumn,letterpaper]{article}

\usepackage{wacv}
\usepackage{times}
\usepackage{epsfig}
\usepackage{graphicx}
\usepackage{amsmath}
\usepackage{amssymb}

\usepackage{amsfonts}
\usepackage[T1]{fontenc}
\usepackage{enumitem}
\usepackage{xcolor}
\usepackage{fancyvrb}
\usepackage{soul}
\usepackage{multirow}
\usepackage{wrapfig}
\usepackage{bm}
\usepackage{lipsum}
\usepackage{booktabs}
\usepackage[accsupp]{axessibility}  

\def\wacvPaperID{0859} 

\wacvfinalcopy 

\ifwacvfinal
\fi

\ifwacvfinal
\usepackage[breaklinks=true,bookmarks=false]{hyperref}
\else
\usepackage[pagebackref=true,breaklinks=true,colorlinks,bookmarks=false]{hyperref}
\fi

\begin{document}

\title{Sandwich Batch Normalization: \\ A Drop-In Replacement for Feature Distribution Heterogeneity}

\author{Xinyu Gong \quad Wuyang Chen \quad Tianlong Chen \quad Zhangyang Wang \\
Department of Electrical and Computer Engineering, the University of Texas at Austin \\
{\tt\small \{xinyu.gong, wuyang.chen, tianlong.chen, atlaswang\}@utexas.edu}}

\maketitle
\thispagestyle{empty}
\begin{abstract}
We present Sandwich Batch Normalization (\textbf{SaBN}), a frustratingly easy improvement of Batch Normalization (BN) with only a few lines of code changes. SaBN is motivated by addressing the inherent \textit{feature distribution heterogeneity} that one can be identified in many tasks, which can arise from data heterogeneity (multiple input domains) or model heterogeneity (dynamic architectures, model conditioning, etc.). Our SaBN factorizes the BN affine layer into one shared \textit{sandwich affine} layer, cascaded by several parallel \textit{independent affine} layers.
Concrete analysis reveals that, during optimization, SaBN promotes balanced gradient norms while still preserving diverse gradient directions -- a property that many application tasks seem to favor. We demonstrate the prevailing effectiveness of SaBN as a \underline{drop-in replacement in four tasks}: \textit{conditional image generation}, \textit{neural architecture search} (NAS), \textit{adversarial training}, and \textit{arbitrary style transfer}. Leveraging SaBN immediately achieves better Inception Score and FID on CIFAR-10 and ImageNet conditional image generation with three state-of-the-art GANs; boosts the performance of a state-of-the-art weight-sharing NAS algorithm significantly on NAS-Bench-201; substantially improves the robust and standard accuracies for adversarial defense; and produces superior arbitrary stylized results. We also provide visualizations and analysis to help understand why SaBN works. 
Codes are available at: \url{https://github.com/VITA-Group/Sandwich-Batch-Normalization}.
\end{abstract} 
\vspace{-1em}
\section{Introduction}
\begin{figure*}[ht!]
\centering
 \includegraphics[width=\linewidth]{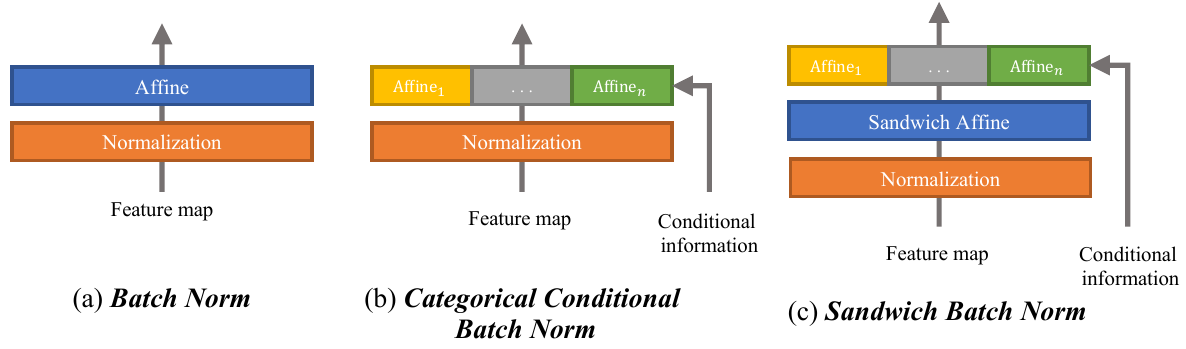}
   \caption{Illustration of (a) vanilla batch normalization (\textbf{BN}), composed of one normalization layer and one affine layer; (b) \textbf{Categorical Conditional BN}, composed of one normalization layer following a set of independent affine layers to intake conditional information;  (c) \textbf{Sandwich BN}, sequentially composed of one normalization layer, one shared sandwich affine layer, and a set of independent affine layers.}
\label{fig:architect}
\end{figure*}

\vspace{-0.5em}
This paper presents a simple, light-weight, and easy-to-implement modification of Batch Normalization (BN) \cite{ioffe2015batch}, motivated by various observations \cite{zajkac2019split,deecke2018mode,xie2019adversarial,xie2019intriguing} drawn from several applications, that \textit{BN has troubles standardizing hidden features with a heterogeneous, multi-modal distribution}. We call this phenomenon \textit{feature distribution heterogeneity}. Such heterogeneity of hidden features could arise from multiple causes, often application-dependent: \vspace{-0.3em}
\begin{itemize}
    \item One straightforward cause is the input \textit{data heterogeneity}. For example, when training a deep network on a diverse set of visual domains, that possess significantly different statistics, BN is found to be ineffective in normalizing the activations with only a single mean and variance \cite{deecke2018mode}, and often needs to be re-set or adapted \cite{li2016revisiting}.\vspace{-0.3em}
    \item Another intrinsic cause could arise from the \textit{model heterogeneity}, i.e., when the training is, or could be equivalently viewed as, on a set of different models. For instance, in neural architecture search (NAS) using weight sharing \cite{liu2018darts,dong2019searching}, training the supernet during the search phase could be considered as training a large set of sub-models (with many overlapped weights) simultaneously. As another example, for conditional image generation \cite{miyato2018spectral}, the generative model could be treated as a set of category-specific sub-models packed together, one of which would be ``activated'' by the conditional input each time. \vspace{-0.3em}
\end{itemize}
The vanilla BN (Figure \ref{fig:architect} (a)) fails to perform well when there is data or model heterogeneity. Recent trends split the affine layer into multiple ones and leverage input signals to modulate or select among them \cite{de2017modulating,deecke2018mode} (Figure \ref{fig:architect} (b)); or even utilize several independent BNs to address such disparity \cite{zajkac2019split,xie2019adversarial,xie2019intriguing,yu2018slimmable}. While those relaxations alleviate the data or model heterogeneity, we suggest that they might be  \textit{``too loose''} in the normalization or regularization effects. 

Let us take the conditional image generation task of GANs as a concrete motivating example to illustrate our rationale. A GAN model is trained with an image dataset containing various image classes, tending to capture the distribution of real samples to produce similar image examples. As a helpful remedy for the generator to encode class-specific information, Categorical Conditional Batch Normalization (CCBN), is widely used in the conditional image generation~\cite{brock2018large, miyato2018spectral, miyato2018cgans}. It is composed of a normalization layer and a number of following separate affine layers for different image classes, which allows class-specific information to be encoded separately, thus enables the generator to generate more vivid examples.

But what might be missing? Unfortunately, using separate affines ignores one important fact that different image classes, while being different, are \textbf{not totally independent}. Considering that images from the same dataset share some common characteristic  (e.g., the object-centric bias for CIFAR images), it is convincing to hypothesize the different classes to be largely overlapped at least (i.e., they still share some hidden features despite the different statistics). To put it simply: while it is oversimplified to normalize the different classes as ``the same one'', it is also unfair and unnecessary to treat them as ``totally disparate''. 

More application examples can be found that all share this important structural feature prior. \cite{liu2018darts,dong2019searching,yu2018slimmable} train a large variety of child models, constituting model heterogeneity; but most child architectures inevitably have many weights in common since they are sampled from the same supernet. Similarly, in adversarial training, the model is trained by a mixture of the original training set (``\textit{clean examples}'') and its attacked counterpart with some small perturbations applied (``\textit{adversarial examples}''). But the clean examples and adversarial examples could be largely overlapped, considering that all adversarial images are generated by perturbing clean counterparts only minimally.

\underline{\textbf{Our Contributions:}} Recognizing the need to address feature normalization with ``harmony in diversity'', we propose a new \textbf{SaBN} as illustrated in Fig \ref{fig:architect} (c). SaBN modifies BN in an embarrassingly simple way: it is equipped with two cascaded affine layers: a shared unconditional \textit{sandwich affine} layer, followed by a set of independent affine layers that can be conditioned. Compared to CCBN, the new sandwich affine layer is designed to inject an inductive bias, that all re-scaling transformations will have a shared factor, indicating the commodity. 

We then dive into a detailed analysis of why SaBN shows to be effective, and illustrate that during optimization, SaBN promotes \textbf{balanced gradient norms} (leading to more fair learning paces among heterogeneous classes and features), while still preserving \textbf{diverse gradient directions} (leading to each class leaning towards discriminative feature clusters): a favorable inductive bias by many applications.
 
Experiments on the applications of conditional image generation and NAS demonstrate that SaBN addresses the \textit{model heterogeneity} issue elegantly, improving generation quality in GAN and the search performance in NAS in a plug-and-play fashion.
To better address the \textit{data heterogeneity} altogether, SaBN could further integrate the idea of split/auxiliary BNs \cite{zajkac2019split,xie2019adversarial,xie2019intriguing,yu2018slimmable}, to decompose the normalization layer into multiple parallel ones. That yields the new variant called \textbf{SaAuxBN}, demonstrated by the example of adversarial training. Lastly, we extend the idea of SaBN to Adaptive Instance Normalization (AdaIN) \cite{huang2017arbitrary} and show the resulting \textbf{SaAdaIN} to improve arbitrary style transfer.

\section{Related Work}

\subsection{Normalization in Deep Learning}
 Batch Normalization (BN)~\cite{ioffe2015batch} made critical contributions to training deep convolutional networks and nowadays becomes a cornerstone of the latter for numerous tasks. BN normalizes the input mini-batch of samples by the mean and variance, and then re-scales them with learnable affine parameters. The success of BNs was initially attributed to overcoming internal covariate shift \cite{ioffe2015batch}, but later on raises many open discussions on its effect of improving landscape smoothness \cite{santurkar2018does}; enabling larger learning rates \cite{bjorck2018understanding} and reducing gradient sensitivity \cite{arora2018theoretical}; preserving the rank of pre-activation weight matrices \cite{daneshmand2020theoretical}; decoupling feature length and direction \cite{kohler2018exponential}; capturing domain-specific artifacts \cite{li2016revisiting}; reducing BN's dependency on batch size~\cite{ioffe2017batch, singh2020filter}; preventing elimination singularities \cite{qiao2019rethinking}; and even characterizing an important portion of network expressivity~\cite{frankle2020training}.
 
 Inspired by BN, a number of task-specific modifications exploit different normalization axes, such as Instance Normalization (IN)  \cite{ulyanov2016instance} for style transfer; Layer Normalization (LN) \cite{ba2016layer} for recurrent networks;  Group Normalization (GN) \cite{wu2018group} for tackling small batch sizes;  StochNorm \cite{kou2020stochastic} for fine-tuning; Passport-aware Normalization \cite{zhang2020passport} for model IP protection; and \cite{li2019positional, wang2020attentive, zheng2020learning} for image generation.

Several normalization variants have been proposed by modulating BN parameters, mostly the affine layer (mean and variance), to improve the controlling flexibility for more sophisticated usages. For example, Harm \etal \cite{de2017modulating} presented Conditional BN, whose affine parameters are generated as a function of the input. Similarly, Conditional IN \cite{dumoulin2016learned} assigned each style with independent IN affine parameters. In \cite{miyato2018spectral}, the authors developed Categorical Conditional BN for conditional GAN image generation, where each generated class has its independent affine parameters. Huang \& Belongie \cite{huang2017arbitrary} presented Adaptive IN (AdaIN), which used the mean and variance of style image to replace the original affine parameter, achieving arbitrary style transfer. Spatial adaptivity \cite{park2019semantic} and channel attention \cite{li2019attentive} managed to modulate BN with higher complexities. 

A few latest works investigate to use multiple normalization layers instead of one in BN. \cite{deecke2018mode} developed mode normalization by employing a mixture-of-experts to separate incoming data into several modes and separately normalizing each mode. \cite{zajkac2019split} used two separate BNs to address the domain shift between labeled and unlabeled data in semi-supervised learning. Very recently, \cite{xie2019intriguing,xie2019adversarial} revealed the two-domain issue in adversarial training and find improvements by using two separate BNs (AuxBN).

\subsection{Brief Backgrounds for Related Applications}
We use four important applications as testbeds. All of them appear to be oversimplified by using the vanilla BN, where the feature homogeneity and heterogeneity are not properly handled. We briefly introduce them below, and will concretely illustrate where the heterogeneity comes from and how our methods deal with it in Sec. \ref{sec:SaBN}.

\paragraph{Generative Adversarial Network}
GAN has been prevailing since its origin \cite{goodfellow2014generative} for image generation. Many efforts have been made to improve GANs, such as modifying loss function \cite{arjovsky2017wasserstein, gulrajani2017improved,jolicoeur2018relativistic}, improving network architecture \cite{zhang2018self,karras2019style,gong2019autogan,chen2020distilling} and adjusting training procedure \cite{karras2017progressive}. Recent works also tried to improve the generated image quality by proposing new normalization modules, such as Categorical Conditional BN and spectral normalization~\cite{miyato2018spectral}. 

\paragraph{Neural Architecture Search (NAS)} The goal of NAS is to automatically search for an optimal model architecture for the given task and dataset.  It was first proposed in \cite{zoph2016neural} where a reinforcement learning algorithm iteratively samples, trains and evaluates candidate models from the search space. Due to its prohibitive time cost, the weight-sharing mechanism was introduced \cite{pham2018efficient} and becomes a popular strategy \cite{liu2018darts}. However, weight-sharing causes performance deterioration due to unfair training \cite{chu2019fairnas}, motivating other alternatives to accelerate NAS \cite{yang2021hournas,chen2020neural,mellor2021neural}. Besides, a few NAS benchmarks \cite{ying2019bench,dong2020bench,zela2020bench} were recently released, with ground-truth accuracy for candidate models pre-recorded, enabling researchers to evaluate the performance of the search method more easily \cite{xu2021renas,chen2021neural,wu2021weak}.

\vspace{-5pt}
\paragraph{Adversarial Robustness}
Deep networks are notorious for the vulnerability to adversarial attacks \cite{goodfellow2014explaining}. In order to enhance adversarial robustness, countless defense approaches have been proposed \cite{dhillon2018stochastic,papernot2017extending,xu2017feature,meng2017magnet,liao2018defense,madry2017towards,dong2020adversarially}. Among them, adversarial training (AT) \cite{madry2017towards} is arguably the strongest, which trains the model over a mixture of clean and perturbed data. The normalization in AT had not been studied in-depth until the pioneering work \cite{xie2019adversarial} introduced an auxiliary batch norm (AuxBN) to improve the clean image recognition accuracy.

\vspace{-5pt}
\paragraph{Neural Style Transfer}
Style transfer~\cite{gatys2016image} generates a stylized image, by combining the content of one image with the style of another. Various improvements are made on the normalization methods \cite{chen2018gated}. \cite{ulyanov2016instance} proposed Instance Normalization (IN), improving the stylized quality of generated images. Conditional Instance Normalization (CIN) \cite{dumoulin2016learned} and Adaptive Instance Normalization (AdaIN) \cite{huang2017arbitrary} enable a single network to perform multiple/arbitrary style transfer.
\section{Sandwich Batch Normalization\label{sec:SaBN}}
\begin{figure*}[ht!]
    \centering
    \includegraphics[width=1.0\linewidth]{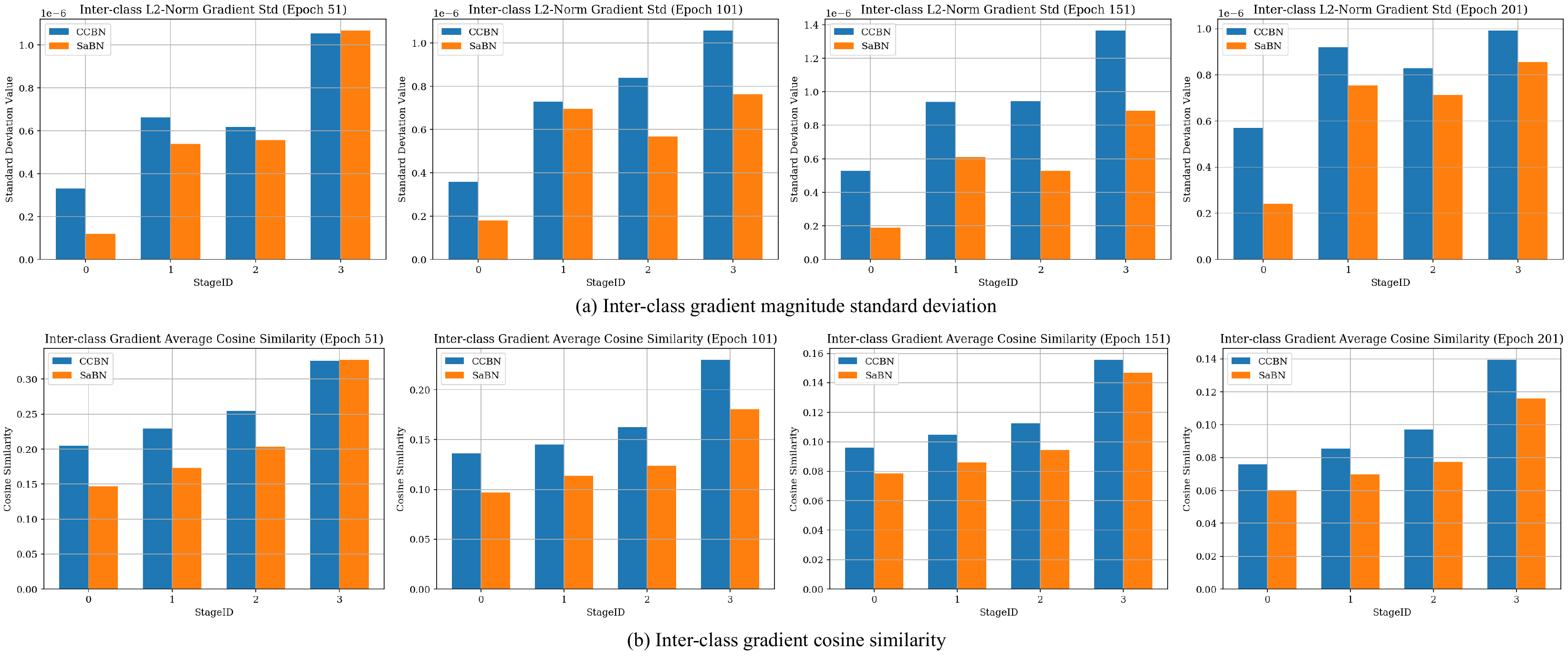}
    \caption{\textbf{The visualization of standard deviations of gradient magnitudes} (the lower the better, i.e., more balanced optimization paces) and cosine similarity across different classes (the lower the better, i.e., more diverse features learned). The x-axis of each plot denotes the depth of generator network, where each generator is composed of four stages of convolution blocks.}
    \label{fig:gan_gradient}
\end{figure*}
Given the input feature $\mathbf{x} \in \mathbb{R}^{N\times C\times H\times W}$ ($N$ denotes the batch size, $C$ the channel number, $H$ height and $W$ width), the vanilla \textbf{Batch Normalization} (\textbf{BN}) works as:
\begin{equation}
\mathbf{h} = \boldsymbol{\gamma} (\frac{\mathbf{x} - \mu(\mathbf{x})}{\sigma(\mathbf{x})}) + \boldsymbol{\beta},
\end{equation}
where $\mu(\mathbf{x})$ and $\sigma(\mathbf{x})$ are the running estimates (or batch statistics) of input $\mathbf{x}$'s mean and variance along ($N$, $H$, $W$) dimensions. $\boldsymbol{\gamma}$ and $\boldsymbol{\beta}$ are the learnable parameters of the affine layer, and both are of shape $C$.
However, the vanilla BN only has a single re-scaling transform and will treat any latent feature from one single distribution.

As an improved variant, \textbf{Categorical Conditional BN} (\textbf{CCBN}) \cite{miyato2018spectral} is proposed to remedy the heterogeneity issue in the task of conditional image generation, boosting the quality of generated images. CCBN has a set of independent affine layers, whose activation is conditioned by the input domain index and each affine layer is learned to capture the class-specific information. It can be expressed as: 
\begin{equation}
\mathbf{h} = \boldsymbol{\gamma}_i (\frac{\mathbf{x} - \mu(\mathbf{x})}{\sigma(\mathbf{x})}) + \boldsymbol{\beta}_i, i = 1, ..., C ,
\end{equation}
where $\boldsymbol{\gamma}_i$ and $\boldsymbol{\beta}_i$ are parameters of the $i$-th affine layer. Concretely, $i$ is the expected output class in the image generation task \cite{miyato2018cgans}.  
However, we argue that this ``separate/split'' modification might cause imbalanced learning for different classes. Since the training data from each class might vary a lot (different number of examples, complicated/simple textures, large/small inner-class variation, etc.), different individual affine layers might have significantly diverged convergence speed, impeding the proper training of the whole network. Dominant classes will introduce stronger inductive biases on convolutional layers than minor classes.

To better handle the imbalance, we present \textbf{Sandwich BN} (\textbf{SaBN}), that is equipped with a shared sandwich affine layer and a set of independent affine layers:
\begin{equation}\label{eq:sabn}
\mathbf{h} = \boldsymbol{\gamma}_i(\boldsymbol{\gamma}_{sa} (\frac{\mathbf{x} - \mu(\mathbf{x})}{\sigma(\mathbf{x})}) + \boldsymbol{\beta}_{sa})  + \boldsymbol{\beta}_i,  i = 1, ..., C .
\end{equation}
As depicted in Fig. \ref{fig:architect} (d), $\boldsymbol{\gamma}_{sa}$ and $\boldsymbol{\beta}_{sa}$ denote the new sandwich affine layer, while $\boldsymbol{\gamma}_i$ and $\boldsymbol{\beta}_i$ are the $i$-th affine parameters, conditioned on categorical inputs. Implementation-wise, SaBN only takes \textbf{a few lines of code changes} over BN: please see the \textbf{supplement} for pseudo codes.

\subsection{Why SaBN meaningfully works?}
One might be curious about the effectiveness of SaBN, since at the inference time, the shared sandwich affine layer can be multiplied/merged into the independent affine layers, making the inference form of SaBN completely identical to CCBN. So where is its real advantage? 

By the analysis below, we argue that: SaBN provides \textit{a favorable inductive bias for optimization}. During training, we observe that SaBN promotes \textbf{balanced gradient norms} (leading to more fair learning paces among heterogeneous classes and features), while still preserving \textbf{diverse gradient directions} (leading to each class leaning towards discriminative feature clusters).



We take the training of the conditional image generation task as an example. As one of the state-of-the-art GANs, SNGAN \cite{miyato2018spectral} successfully generates high-quality images in the conditional image generation task with CCBN. Intuitively, it uses independent affine layers to disentangle the image generation of different classes.
We consider analyzing the following two models: 1) SNGAN (equipped with CCBN originally); 2) SNGAN-SaBN (simply replaces CCBN with our SaBN). Both of them are trained on the same subset of ImageNet, with the same training recipe (see Sec. \ref{sec:gan} for details). Here we analyze the difference of inter-class gradients of the generator in two aspects: (1) gradient magnitude and (2) gradient direction.

\textbf{SaBN Encourages Balanced Gradient Magnitudes.}
For the first aspect, we analyze the standard deviation of the $l_2$-norm (magnitude) of generator's gradients among different classes in Fig. \ref{fig:gan_gradient} (a). Concretely, we take the gradients from weights of convolution layers, which are right before the normalization modules. We observe that the gradient norms from SNGAN-SaBN mostly have lower standard deviations, indicating that the gradient magnitudes of different classes in SNGAN-SaBN are more balanced. A balanced distribution of gradient magnitudes is found to be preferred for model training, avoiding some features dominating the others, and facilitating the optimization of all sub-tasks at similar paces~\cite{yu2020gradient}.


\textbf{SaBN Preserves Diversity of Gradient Directions.}
We then visualize the averaged cosine similarity of generator's gradients from different classes during training in Fig. \ref{fig:gan_gradient} (b). Specifically, we define the inter-class gradient similarity $g_{\text{inter}}$, which aims to measure the divergence of gradients from different input class labels $y$ and averaged over latent vectors ($z$):
\begin{equation}\label{eq:cos_sim}
g^l_{\text{inter}} = \frac{1}{m} \sum_{i=0}^{m-1}  \mathcal{C}\left(\left.\nabla_{\theta^l} \mathcal{L}(G(z_i, y_j))\right|_{j=0} ^{n-1}\right) \\ 
\end{equation}
Here the generator is denoted by $G$, and $\nabla_{\theta^l} \mathcal{L}(G(z_i, y_j))$ represents the gradients on convolution layers of the $l$-th stage of the generator (i.e., the derivative of loss $\mathcal{L}$ with respect to parameters in $l$-th stage $\theta^l$; we omit the discriminator here). $m$ and $n$ are the total number of latent vectors $z$ and class labels $y$, respectively. Function $\mathcal{C}$ calculates the averaged pair-wise cosine similarity of inputs:
\begin{equation}\label{eq:cosine}
\mathcal{C} \left(\mathbf{v}_i|_{i=1} ^{N}\right) = \frac{1}{N(N-1)} \sum_{i=1}^{N} \sum_{j=1, j \neq i}^{N} \frac{\mathbf{v}_i \cdot \mathbf{v}_j}{\|\mathbf{v}_i\|\|\mathbf{v}_j\|}.
\end{equation}
We can see that SNGAN-SaBN has lower $g_{\text{inter}}$, indicating the gradients from different classes are more diverse in their directions. This enables the generator to capture richer discriminative information among different classes and contribute to a visually more diverse generation. 

\begin{figure}[t!]
\begin{center}
 \includegraphics[width=\linewidth]{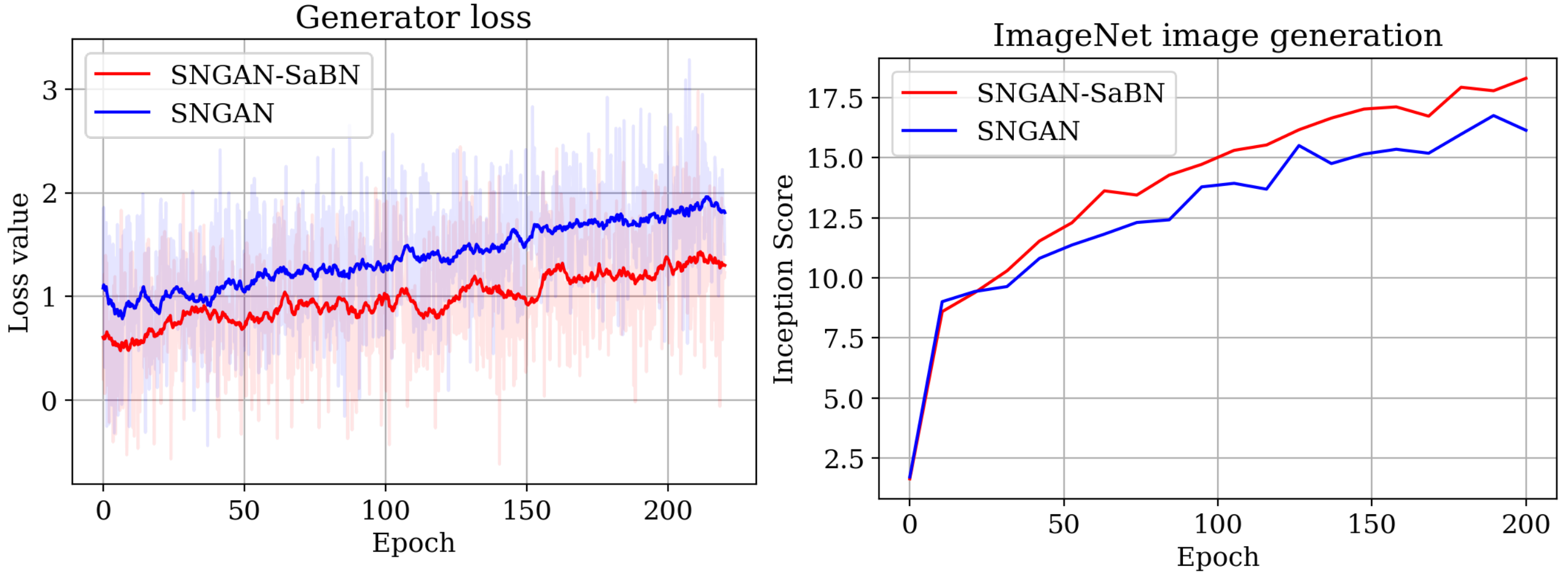}
\end{center}
   \caption{\textbf{(left)} The generator loss $L_{G}$ of SNGAN and SNGAN-SaBN during training phase (ImageNet). (\textbf{right}) The Inception Score curves during training.}
\label{fig:g_loss}
\end{figure}
\textbf{Summary:} The above two characteristics brought by SaBN, i.e., overall more balanced gradient norms, and inter-class more diverse gradient directions, together draw the big picture why SaBN facilitates the optimization especially in the presence of heterogeneous or multi-domain features. Fig. \ref{fig:g_loss} visualizes GAN training with and without SaBN: SNGAN with SaBN can achieve much lower generator loss $L_{G}$ (Eq. \ref{eq:gan_loss}) than the original SNGAN (Fig. \ref{fig:g_loss} left), and the generation quality of the former consistently outperforms the latter (by Inception Score~\cite{salimans2016improved}, Fig. \ref{fig:g_loss} right).


\section{Experiments}
Sandwich Batch Normalization is an effective plug-and-play module. In this section, we present the experiment results of naively applying it into two different tasks: conditional image generation and neural architecture search. 

\subsection{Conditional Image Generation with SaBN} \label{sec:gan}
Following the discussion in the previous section, we present detailed settings and main results on the conditional image generation task using SaBN in this section. We choose three representative GAN models, SNGAN, BigGAN \cite{brock2018large} and AutoGAN-top1 \cite{gong2019autogan}, as our baselines. The generator of SNGAN and BigGAN are equipped with CCBN originally. AutoGAN-top1 does not have any normalization layer and is designed for unconditional image generation, thus we manually insert CCBN into its generator to adapt it to the conditional image generation task. We then construct SNGAN-SaBN, BigGAN-SaBN, and AutoGAN-top1-SaBN, by simply replacing all CCBN in the above baselines with our proposed SaBN. 

GAN models in our paper are all trained with the hinge version adversarial loss~\cite{miyato2018spectral, brock2018large}:
\begin{equation} \label{eq:gan_loss}
\begin{aligned}
L_{D}=&-\mathbb{E}_{(x, y) \sim p_{\text {data }}}[\min (0,-1+D(x, y))] \\
&-\mathbb{E}_{z \sim p_{z}, y \sim p_{\text {data }}}[\min (0,-1-D(G(z, y), y))], \\
L_{G}=&-\mathbb{E}_{z \sim p_{z}, y \sim p_{\text {data }}} D(G(z, y), y),
\end{aligned}
\end{equation}
where $L_{D}$ and $L_{G}$ denote discriminator loss and generator loss respectively.

\footnotetext[1]{Results obtained by using the \href{https://github.com/ajbrock/BigGAN-PyTorch}{author's officially unofficial PyTorch BigGAN implementation}.}

\begin{table}[t!]
\caption{Best \textbf{Inception Scores} (``IS'', $\uparrow$) and \textbf{FIDs} ($\downarrow$) achieved by conditional SNGAN, BigGAN, and AutoGAN-top1, using CCBN and SaBN on CIFAR-10 and ImageNet (dogs \& cats).}
\label{tab:gan}
\footnotesize
\centering
\resizebox{\linewidth}{!}{
\begin{tabular}{ l|cc|cc }
\toprule
& \multicolumn{2}{ c }{CIFAR-10} &  \multicolumn{2}{ c }{ImageNet (dogs \& cats)}\\ \midrule
Model & IS & FID & IS & FID \\ \midrule
AutoGAN-top1 & $8.43$ & $10.51$ & - & - \\
BigGAN & $8.91^1$ & $8.57^1$ &  - & -  \\
SNGAN & $8.76$ & $10.18$ & $16.75$ & $79.14 $ \\\midrule
\textbf{AutoGAN-top1-SaBN} & $8.72 (+0.29)$ & $9.11 (-1.40)$ & - & - \\
\textbf{BigGAN-SaBN} & $ 9.01 (+0.10)$ & $ 8.03 (-0.54)$ & - & - \\
\textbf{SNGAN-SaBN}& $8.89 (+0.13)$ & $8.97 (-1.21)$ & $18.31 (+1.56)$ & $60.38 (-18.76)$ \\
\bottomrule
\end{tabular}
}
\end{table} 
We test all the above models on CIFAR-10 dataset \cite{krizhevsky2009learning} (10 categories, resolution $32 \times 32$). Furthermore, we test SNGAN and SNGAN-SaBN on high-resolution conditional image generation task with ImageNet \cite{deng2009imagenet}, using the subset of all 143 classes belonging to the dog and cat super-classes, cropped to resolution $128 \times 128$) following \cite{miyato2018spectral}'s setting. Inception Score \cite{salimans2016improved} (the higher the better) and FID \cite{heusel2017gans} (the lower the better) are adopted as evaluation metrics. We summarize the best performance the models have achieved during training into Table \ref{tab:gan}. We find that SaBN can consistently boost the generative quality of all three baseline GAN models, which demonstrates the effectiveness of the injected shared sandwich affine layer.
\begin{figure}[t!]
\begin{center}
 \includegraphics[width=\linewidth]{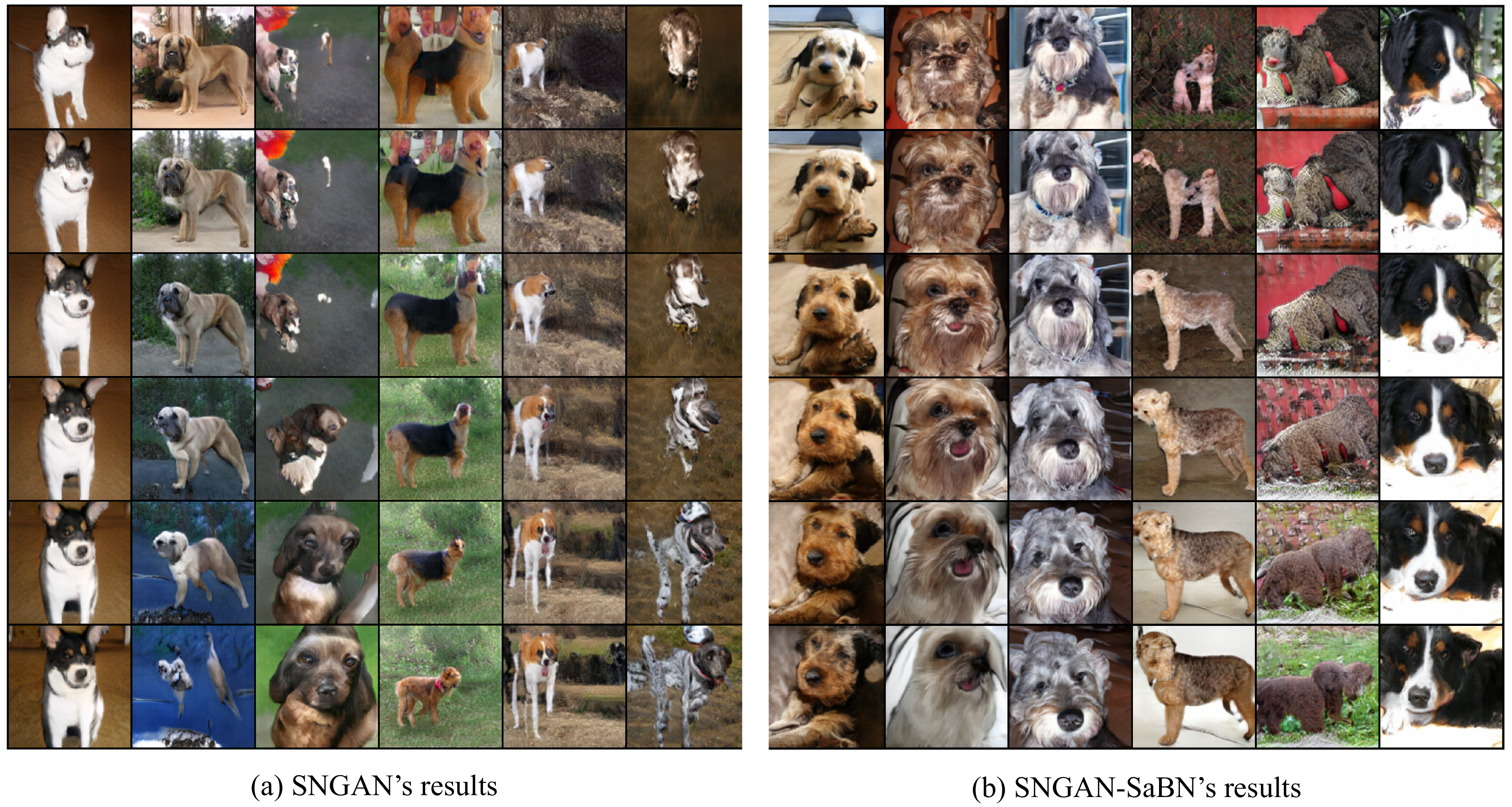}
\end{center}
   \caption{\textbf{The image generation results of SNGAN and SNGAN-SaBN on ImageNet.} Each column is corresponding to a specific image class. Images are chosen without cherry-pick.}
\label{fig:gan_vis}
\end{figure}
We also provide visualization results of generated images in Fig. \ref{fig:gan_vis}. 
Since images of CIFAR-10 dataset~\cite{krizhevsky2009learning} are too small to tell difference, we only visualize the results on ImageNet~\cite{deng2009imagenet}. Specifically, we compare the generation results of SNGAN and SNGAN-SaBN. The images generated by SNGAN-SaBN are more visual appealing, showing better quality.

\subsection{Architecture Heterogeneity in Neural Architecture Search (NAS)}
Recent NAS works formulate the search space as a weight-sharing supernet that contains all candidate operations and architectures, and the goal is to find a sub-network that of the optimal performance. As one of the representative works in NAS, DARTS~\cite{liu2018darts} solves the search problem by assigning each candidate operation a trainable architecture parameter $\alpha$. Model weights $\boldsymbol{\omega}$ and architecture parameters $\boldsymbol{\alpha}$ are optimized to minimize the cross-entropy loss in an alternative fashion. After searching, the final architecture is derived by choosing the operation with the highest $\alpha$ value on each edge.

\begin{figure}[t!]
\begin{center}
 \includegraphics[width=\linewidth]{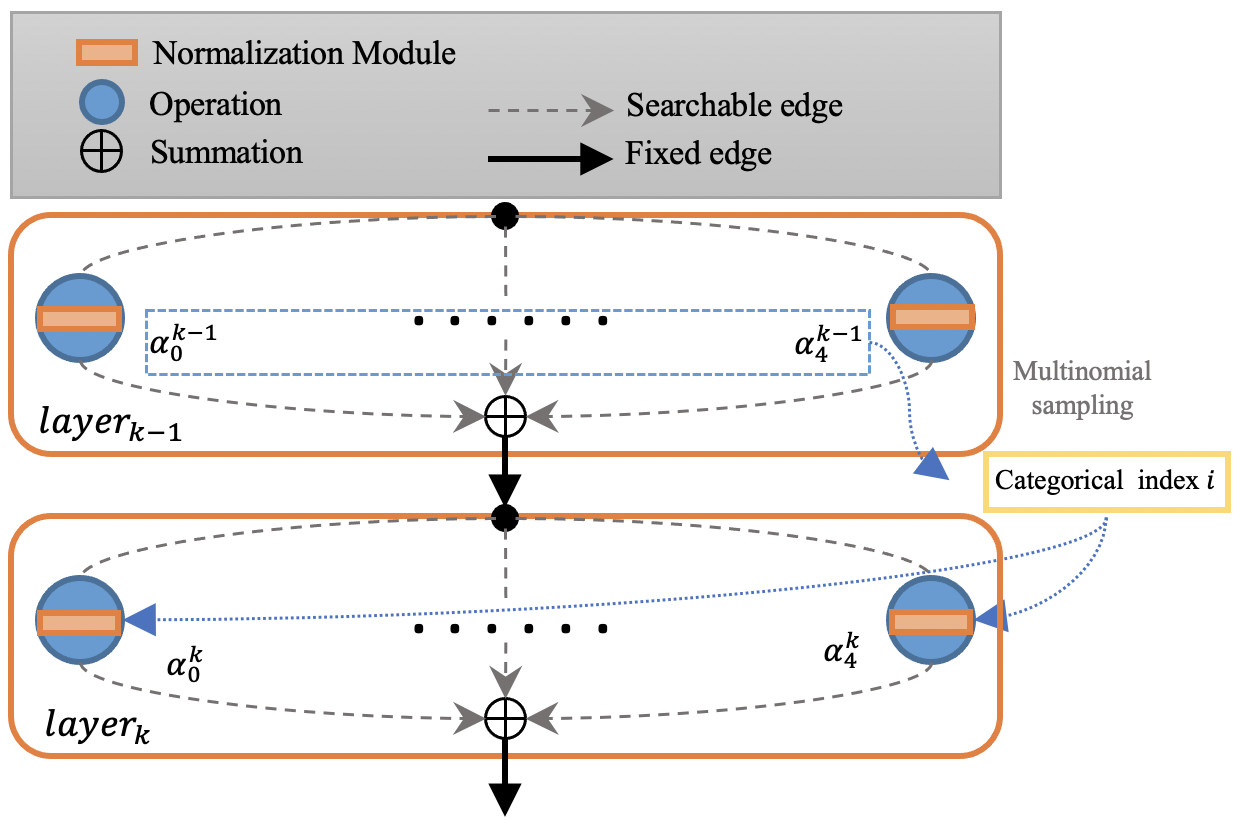}
\end{center}
  \caption{Two consecutive layers in the supernet. By default, a BN is integrated into each parameterized operation in vanilla DARTS. The output of each layer is the sum of all operation paths' output, weighted by their associated architecture parameter $\bm{\alpha}$. 
  }
\label{fig:supernet}
\end{figure}

However, such formulation introduces a strong model heterogeneity. As shown in Fig. \ref{fig:supernet}, the output of each layer is the sum of all operations' output, weighted by associated architecture parameters $\boldsymbol{\alpha}$. Such mixed model heterogeneity could be harmful to the search, making the algorithm hard to distinguish the contribution of each operation. Inspired by the application of CCBN in GANs, we preliminarily attempt to use CCBN disentangling the mixed model heterogeneity from the previous layer, by replacing the BN in each operation path with a CCBN (namely DARTS-CCBN). The total number of affine paths is equal to the number of candidate operations in the previous layer, and the conditional index $i$ of CCBN is obtained by applying a multinomial sampling on the softmax of previous layers’ architecture parameters (shown in  Fig. \ref{fig:supernet}). The search results of the vanilla DARTS and DARTS-CCBN are reported in Tab. \ref{tab:darts}. Compared with vanilla DARTS, DARTS-CCBN does not show consistent improvement w.r.t. the search results. We argue that the brutal “separate/split” modiﬁcation in CCBN might cause imbalanced learning for different operations due to their intrinsic difference, therefore leading to unfair competition among candidate operations. 

To better handle such an unbalanced issue, we consider using SaBN instead of CCBN (DARTS-SaBN). The injected shared sandwich affine layer is designed to balance the learning among different operations, imposing a more fair competition among candidate operations. As shown in Tab. \ref{tab:darts}, we can observe that DARTS-SaBN outperforms the vanilla DARTS and DARTS-CCBN significantly, whose performance is even close to the Bench optimal. The performance gap between DARTS-CCBN and DARTS-SaBN demonstrates the effectiveness of the sandwich affine layer. We further include an additional ablation variant DARTS-affine, which simply enables the affine layer of the BN in DARTS. DARTS-SaBN also outperforms DARTS-affine with a considerable margin, implying the independent conditional affine layers are also important.
\begin{table}[t!]
\caption{\textbf{The ground-truth top-1 accuracy of the final searched architecture on NAS-Bench-201.} DARTS-SaBN achieves the highest accuracy, with the lowest standard deviation. Bench optimal denotes the best test accuracy achievable in NAS-Bench-201.}
\label{tab:darts}
\small
\centering
\resizebox{0.9\linewidth}{!}{
\begin{tabular}{ lcc}
\toprule
Method & CIFAR-100 &  ImageNet16-120 \\ \midrule
DARTS & 44.05 $\pm$ 7.47 & 36.47 $\pm$ 7.06\\
DARTS-affine & 63.46 $\pm$ 2.41 & 37.26 $\pm$ 7.65\\
DARTS-CCBN & 62.16 $\pm$ 2.62 & 31.25 $\pm$ 6.20\\
DARTS-SaBN (ours) & \textbf{71.56 $\pm$ 1.39} & \textbf{45.85 $\pm$ 0.72}\\ \midrule
\textbf{Bench Optimal} & 73.51 & 47.31 \\
\bottomrule
\end{tabular}
}
\end{table}

\section{Extended Applications of Sandwich Batch Normalization}
In this section, we explore the possibility to extend Sandwich Batch Normalization to more tasks with minor modifications. Concretely, we apply two variants of SaBN on adversarial robustness and arbitrary style transfer.
\subsection{Sandwich Auxiliary Batch Norm (SaAuxBN) in Adversarial Robustness} 
AdvProp \cite{xie2019adversarial} successfully utilized adversarial examples to boost network Standard Testing Accuracy (SA) by introducing Auxiliary Batch Norm (AuxBN). The design is quite simple: an additional BN is added in parallel to the original BN, where the original BN (clean branch) takes the clean image as input, while the additional BN (adversarial branch) is fed with only adversarial examples during training. That intuitively disentangles the mixed clean and adversarial distribution (\textbf{data heterogeneity}) into two splits, guaranteeing the normalization statistics and re-scaling are exclusively performed in either domain. The loss function of AdvProp can be formulated as:
\begin{equation} \label{qg:adv_loss}
    L_{\text{total}} = L(f_{\text{clean}}(x_{\text{clean}}), y) + L(f_{\text{adv}}(x_{\text{adv}}), y),
\end{equation}
where $f$ denotes the model and $x$, $y$ denotes the input data, label respectively. $L$ is the cross entropy loss. Therefore, $f_{\text{clean}}$ denotes the model is using the clean branch BN. $x_{\text{adv}}$ is the corresponding adversarial mini-batch generated by the model using adversarial branch BN. For simplicity, we call the two loss terms in Eq. \ref{qg:adv_loss} as \textbf{clean loss} and \textbf{adversarial loss} respectively.

However, one thing missed is that the domains of clean and adversarial images overlap largely, as adversarial images are generated by perturbing clean counterparts minimally. This inspires us to present a novel \textbf{SaAuxBN}, by leveraging domain-specific normalization and affine layers, and also a shared sandwich affine layer for homogeneity preserving. SaAuxBN can be defined as:
\begin{equation}
\mathbf{h} = \boldsymbol{\gamma}_i(\boldsymbol{\gamma}_{sa} (\frac{\mathbf{x} - \mu_i(\mathbf{x})}{\sigma_i(\mathbf{x})}) + \boldsymbol{\beta}_{sa})  + \boldsymbol{\beta}_i, i = 0, 1.
\end{equation}
$\mu_i(\mathbf{x})$ and $\sigma_i(\mathbf{x})$ denote the $i$-th (moving) mean and variance of input, where $i=0$ for adversarial images and $i=1$ for clean images. We use independent normalization layer to decouple the data from two different distributions, i.e., the clean and adversarial. 

We replace AuxBN with SaAuxBN in AdvProp \cite{xie2019adversarial} and find it can further improve SA of the network with its clean branch. The experiments are conducted on CIFAR-10 \cite{krizhevsky2009learning} with ResNet-18 \cite{he2016deep}. For a fair comparison, we follow the settings in \cite{madry2017towards}. In the adversarial training, we adopt $\ell_{\infty}$ based $10$ steps Projected Gradient Descent (PGD) \cite{madry2017towards} with step size $\alpha=\frac{2}{255}$ and maximum perturbation magnitude $\epsilon=\frac{8}{255}$; As for assessing RA, PGD-20 with the same configuration is adopted. The results are presented in Tab. \ref{tab:adv_clean}. 

We further conduct an experiment to test the Standard Testing Accuracy (SA) and Robust Testing Accuracy (RA) of the network using the adversarial branch of AuxBN and SaAuxBN. The comparison results are presented in Tab. \ref{tab:adv}. 
We can see that BN still achieves the highest performance on SA, but falls a lot on RA compared with other methods. Our proposed SaAuxBN is on par with the vanilla BN in terms of SA, while has significantly better results on RA than any other approaches. Compared with SaAuxBN, AuxBN suffers from both worse SA and RA.

\begin{table}[t]
\caption{\textbf{Model performance (SA) using clean branch.}}
\label{tab:adv_clean}
\centering
\resizebox{\linewidth}{!}{
\begin{tabular}{c|cccc}
\toprule
Evaluation & BN & ModeNorm & AuxBN (clean branch)  & SaAuxBN (clean branch)\\ \midrule
Clean (SA) & 84.84 & 83.28 & $94.47 $  & $\textbf{94.62}$\\ 
\bottomrule
\end{tabular}}
\end{table}

\begin{table}[t!]
\caption{\textbf{Performance (SA\&RA) using the adversarial branch.}} 
\label{tab:adv}
\centering
\resizebox{\linewidth}{!}{
\begin{tabular}{c|cccc}
\toprule
Evaluation & BN & ModeNorm & AuxBN (adv branch)  & SaAuxBN (adv branch) \\ \midrule
Clean (SA) & \textbf{84.84} & 83.28 & 83.42 & 84.08 \\ 
PGD-$10$ (RA) & 41.57 & 43.56 & 43.05 &\textbf{44.93} \\ 
PGD-$20$ (RA) & 40.02 & 41.85 & 41.60 &\textbf{43.14} \\ \bottomrule
\end{tabular}}
\end{table}

\begin{figure}[t!]
\begin{center}
 \includegraphics[width=\linewidth]{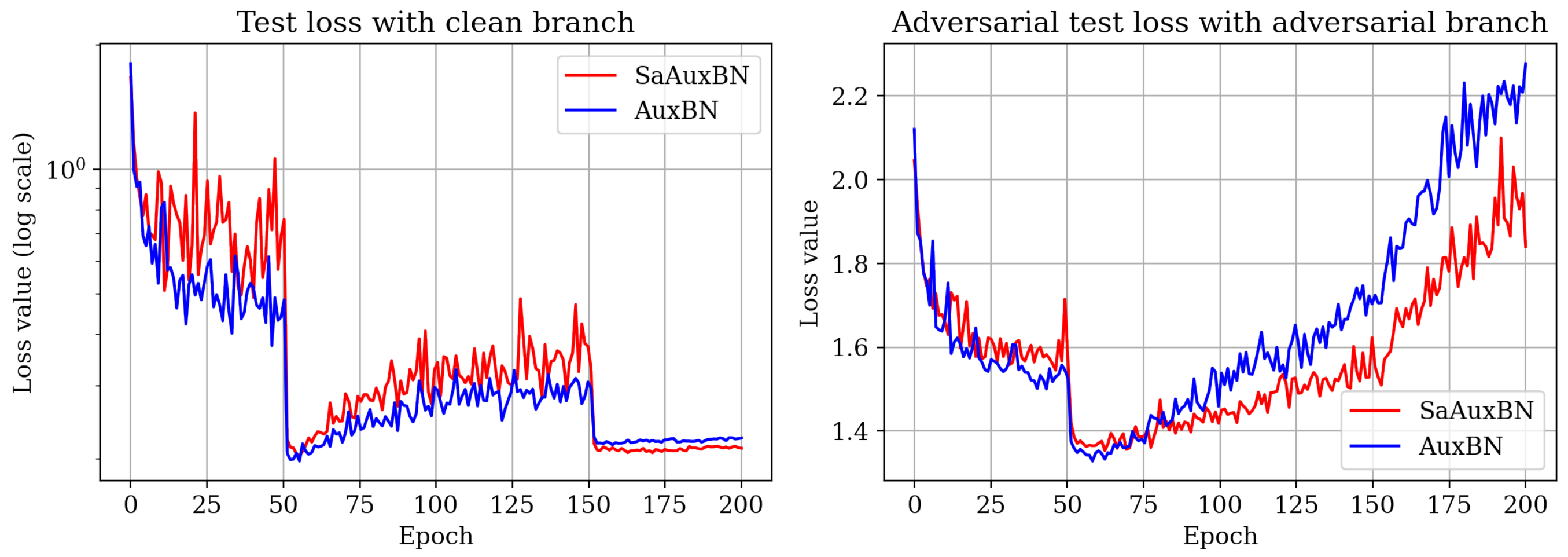}
\end{center}
  \caption{The \textbf{clean loss} $L(f_{\text{clean}}(x_{\text{clean}}), y)$ and \textbf{adversarial loss} $L(f_{\text{adv}}(x_{\text{adv}}), y)$ on testing set.}
\label{fig:robustness_loss}
\end{figure}
We also visualize the testing clean loss and adversarial loss of models with AuxBN and SaAuxBN in Fig. \ref{fig:robustness_loss}, showing that the latter achieves lower value on both.
 
We additionally include ModeNorm \cite{deecke2018mode} as an ablation in our experiments, which was proposed to deal with multi-modal distributions inputs, i.e., data heterogeneity. It shares some similarity with AuxBN as both consider multiple independent norms. ModeNorm achieves fair performance on both SA and RA, while still lower than SaAuxBN. The reason might be the output of ModeNorm is a summation of two features weighted by a set of learned gating functions, which still mixes the statistics from two domains, leading to inferior performance. 

\subsection{Arbitrary Style Transfer with Sandwich Adaptive Instance Normalization (SaAdaIN)}
Huang \& Belongie \cite{huang2017arbitrary} achieves arbitrary style transfer by introducing Adaptive Instance Norm (AdaIN), which is an effective module to encode style information into feature space.  The AdaIN framework is composed of three parts: Encoder, AdaIN, and Decoder. Firstly, the Encoder will extract content features and style features from content and style images. Then, the AdaIN is leveraged to perform style transfer on feature space, producing a stylized content feature. The Decoder is learned to decode the stylized content feature to stylized images. This framework is trained end-to-end with two loss terms, a content loss and a style loss. 
 Concretely, AdaIN firstly performs a normalization on the content feature, then re-scale the normalized content feature with style feature's statistic. It can be formulated as:
 \begin{equation}
 \mathbf{h} = \sigma(\mathbf{y}) (\frac{\mathbf{x} - \mu(\mathbf{x})}{\sigma(\mathbf{x})}) + \mu(\mathbf{y}) ,
 \end{equation}
where $\mathbf{y}$ is the style input, $\mathbf{x}$ is the content input. Note that $\mu$ and $\sigma$ here are quite different from BN, which are performed along the spatial axes ($H, W$) for each sample and each channel. The goal of style transfer is to extract the style information from the style input and render it to the content input. Obviously, style-dependent re-scale may be too loose and might further amplify the intrinsic \textbf{data heterogeneity} brought by the variety of the input content images, undermining the network's ability of maintaining the content information in the output. In order to reduce the \textbf{data heterogeneity}, we propose to insert a shared sandwich affine layer after the normalization, which introduce \textbf{homogeneity} for the style-dependent re-scaling transformation. Hereby, we present \textbf{SaAdaIN}:
\begin{equation}
 \mathbf{h} = \sigma(\mathbf{y})( \boldsymbol{\gamma}_{sa}(\frac{\mathbf{x} - \mu(\mathbf{x})}{\sigma(\mathbf{x})}) +\boldsymbol{\beta}_{sa})+ \mu(\mathbf{y}) ,
\end{equation}
Besides AdaIN, we also include Instance-Level Meta Normalization with Instance Norm  (ILM+IN) proposed by \cite{jia2019instance} as a task-specific comparison baseline. Its style-independent affine is not only conditioned on style information but also controlled by the input feature. 

Our training settings for all models are kept identical with \cite{huang2017arbitrary}.  The network is trained with style loss and content loss. We use the training set of MS-COCO \cite{lin2014microsoft} and WikiArt \cite{wikiart} as the training content and style images dataset, and the validation set of MS-COCO and testing set of WikiArt are used as our validation set.

\begin{figure}[t!]
\begin{center}
 \includegraphics[width=\linewidth]{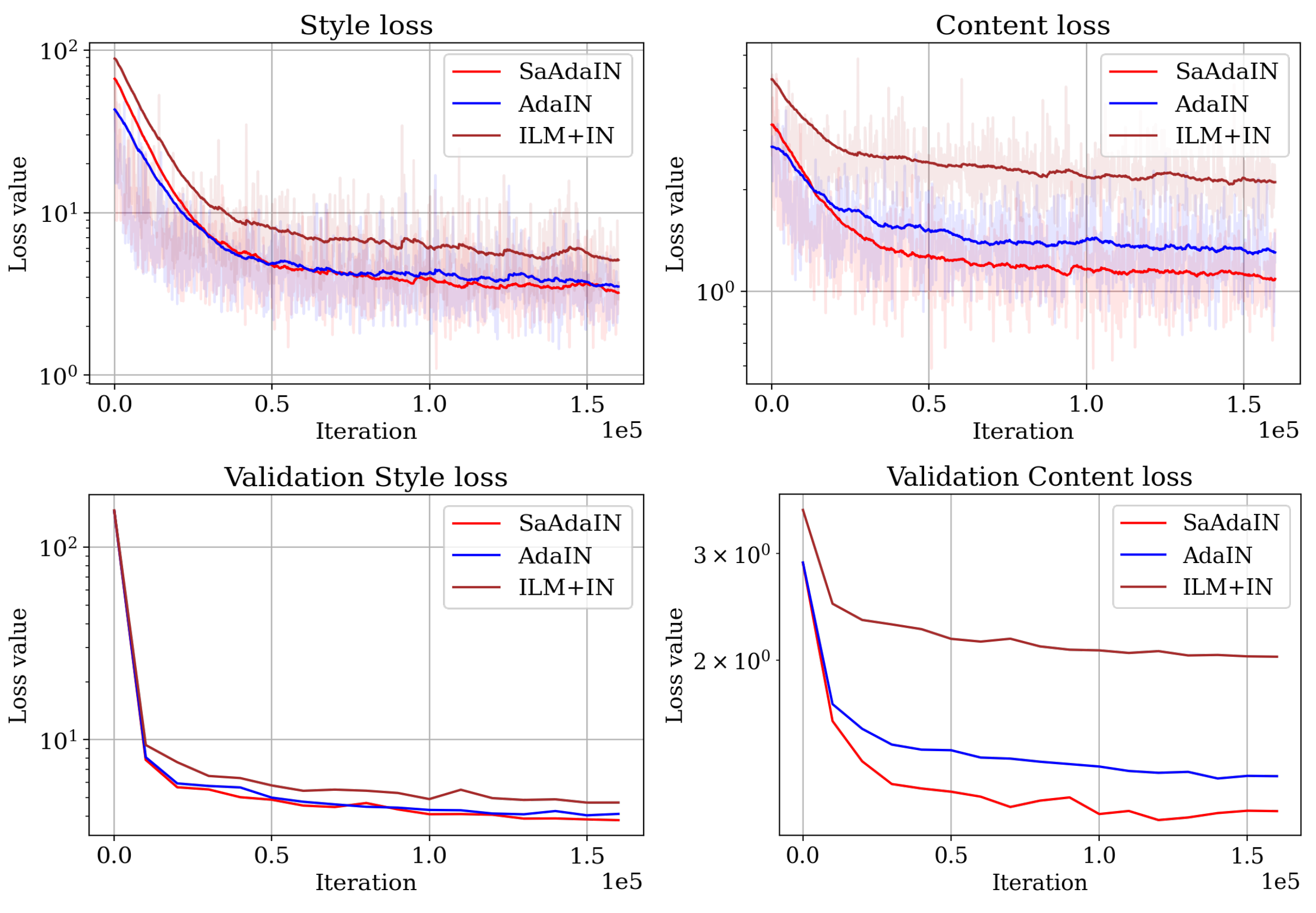}
\end{center}
   \caption{\textbf{The content loss and the style loss of using AdaIN, ILM+IN and SaAdaIN on training and validation set.} In the first row, the noisy shallow-color curves are the original data, and the foreground smoothed curves are obtained via applying exponential moving average on the original data.}
\label{fig:style_curve}
\end{figure}

\begin{figure}[h]
\begin{center}
 \includegraphics[width=1.0\linewidth]{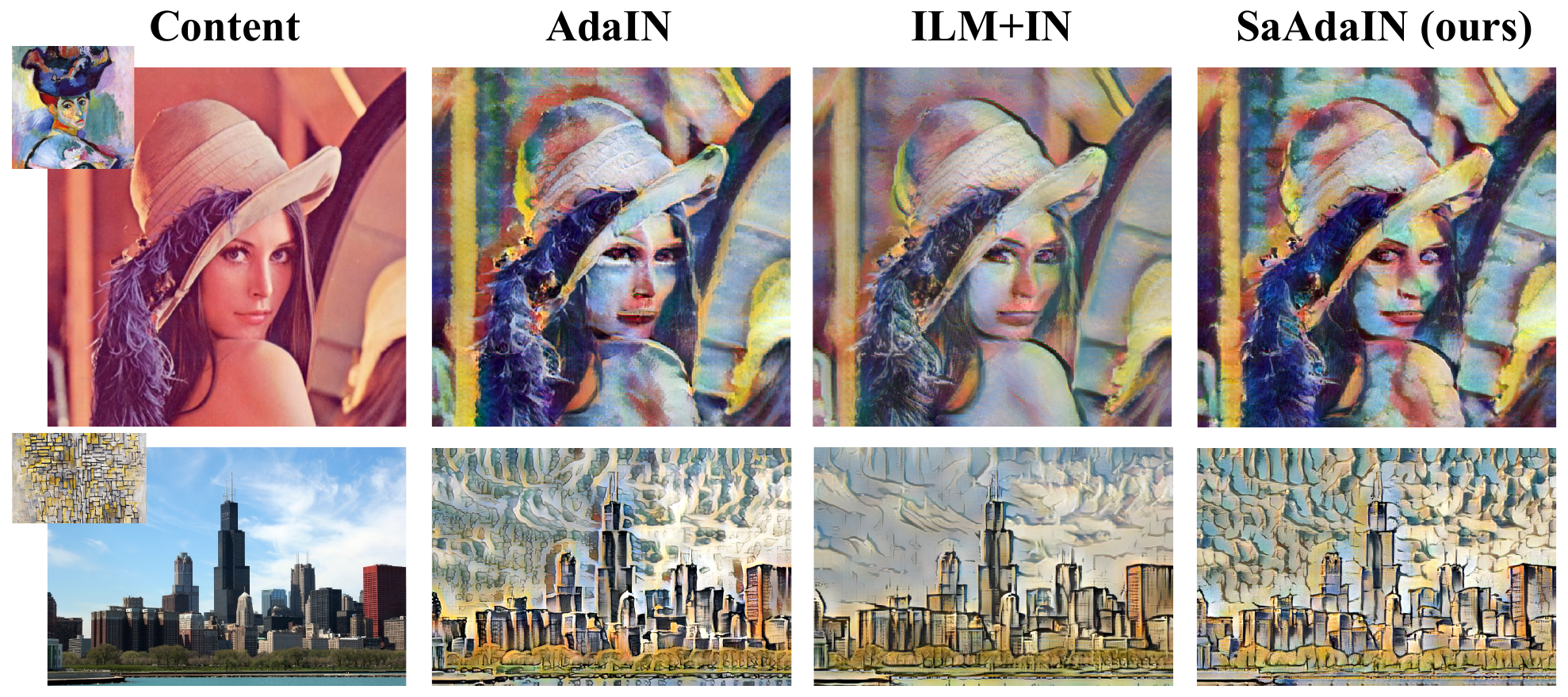}
\end{center}
   \caption{\textbf{The visual results of style transfer.} The images on the top-left corner of each row are the reference style image. An ideally stylized output should be semantically similar to the content image, while naturally incorporate the style information from the referenced style image. }
\label{fig:style_vis}
\end{figure}

We depict the loss curves of the training phase in Fig. \ref{fig:style_curve} (a). We can notice that both the content loss and style loss of the proposed SaAdaIN are lower than that of AdaIN and ILM+IN. This observation implies that the inserted sandwich affine layer makes the optimization easier. In Fig. \ref{fig:style_curve} (b), we show the content and style loss on validation set. In contrast with the AdaIN model, the network with SaAdaIN achieves lower validation content and style loss, indicating the inserted sandwich affine layer also benefits the model's generalizability. The visual results of style transfer are shown in Fig. \ref{fig:style_vis}. 
Compared with AdaIN and ILM-IN, SaBN generates more visual-appealing images.

\section{Conclusion}
We present SaBN and its variants as plug-and-play normalization modules, which are motivated by addressing model \& data heterogeneity issues. We demonstrate their effectiveness on several tasks, including neural architecture search, adversarial robustness, conditional image generation, and arbitrary style transfer. Our future work plans to investigate the performance of SaBN on more applications, such as semi-supervised learning \cite{zajkac2019split}.

\section{Acknowledgment}
 Z. Wang is in part supported by IoBT REIGN subaward \# 088831-18415, US Army Research Laboratory.

{\small
\bibliographystyle{ieee_fullname}
\bibliography{egbib}
}

\end{document}